\documentclass[lettersize,journal]{IEEEtran}
\usepackage{cite}
\usepackage{amsmath,amsfonts}

\usepackage{amsmath,amssymb,amsfonts}
\usepackage{algorithmic}
\usepackage{graphicx}
\usepackage{textcomp}
\usepackage{soul}

\usepackage{hyperref}
\usepackage{makecell}
\usepackage{multirow}
\usepackage{array}
\usepackage{caption}
\usepackage{subcaption}
\usepackage{placeins}
\usepackage{tabularx}
\usepackage{booktabs}
\usepackage{tabularx}
\usepackage{tablefootnote}
\usepackage{color}

\newcolumntype{P}[1]{>{\centering\arraybackslash}p{#1}}

\newcommand\Tstrut{\rule{0pt}{2.0ex}}         % = `top' strut
\newcommand\Bstrut{\rule[-0.3ex]{0pt}{0pt}}

\def\BibTeX{{\rm B\kern-.05em{\sc i\kern-.025em b}\kern-.08em
    T\kern-.1667em\lower.7ex\hbox{E}\kern-.125emX}}
\begin{document}
% \history{Date of publication xxxx 00, 0000, date of current version xxxx 00, 0000.}
% \doi{10.1109/ACCESS.2023.0322000}

\title{Keystroke Verification Challenge (KVC):\\ Biometric and Fairness Benchmark Evaluation}

\author{\IEEEauthorblockN{Giuseppe Stragapede\IEEEauthorrefmark{1}\thanks{Email: \href{mailto:giuseppe.stragapede@uam.es}{giuseppe.stragapede@uam.es}}, 
Ruben Vera-Rodriguez\IEEEauthorrefmark{2}\IEEEauthorrefmark{1}, 
Ruben Tolosana\IEEEauthorrefmark{1}, \\
Aythami Morales\IEEEauthorrefmark{1},
Naser Damer\IEEEauthorrefmark{2},
Julian Fierrez\IEEEauthorrefmark{1} and 
Javier Ortega-Garcia\IEEEauthorrefmark{1}
}

\IEEEauthorblockA{\IEEEauthorrefmark{1}Biometrics and Data Pattern Analytics (BiDA) Lab, Universidad Autonoma de Madrid, Spain}

\IEEEauthorblockA{\IEEEauthorrefmark{2}Fraunhofer Institute for Computer Graphics Research IGD, Darmstadt, Germany}
}

% \author{\IEEEauthorblockN{Giuseppe Stragapede\IEEEauthorrefmark{1}\thanks{Email: \href{mailto:giuseppe.stragapede@uam.es}{giuseppe.stragapede@uam.es}}, 
% Paula Delgado-Santos\IEEEauthorrefmark{2}\IEEEauthorrefmark{1}, 
% Ruben Tolosana\IEEEauthorrefmark{1}, \\
% Ruben Vera-Rodriguez\IEEEauthorrefmark{1},
% Richard Guest\IEEEauthorrefmark{2} and
% Aythami Morales\IEEEauthorrefmark{1}
% }

% \IEEEauthorblockA{\IEEEauthorrefmark{1}Biometrics and Data Pattern Analytics (BiDA) Lab, Universidad Autonoma de Madrid, Spain}

% \IEEEauthorblockA{\IEEEauthorrefmark{2}School of Engineering, University of Kent, United Kingdom}
% }

\maketitle

% \address[2]{Biometrics and Data Pattern Analytics (BiDA) Lab, Universidad Autónoma de Madrid, Spain (email: ruben.vera@uam.es)}
% \address[3]{Biometrics and Data Pattern Analytics (BiDA) Lab, Universidad Autónoma de Madrid, Spain (email: ruben.tolosana@uam.es)}
% \address[4]{Biometrics and Data Pattern Analytics (BiDA) Lab, Universidad Autónoma de Madrid, Spain (email: aythami.morales@uam.es)}

% \tfootnote{This paragraph of the first footnote will contain support
% information, including sponsor and financial support acknowledgment. For
% example, ``This work was supported in part by the U.S. Department of
% Commerce under Grant BS123456.''}

% \markboth
% {Stragapede \headeretal:Keystroke Verification Challenge (KVC): Biometric and Fairness Benchmark Evaluation}
% {Stragapede \headeretal:Keystroke Verification Challenge (KVC): Biometric and Fairness Benchmark Evaluation}

% \corresp{Corresponding author: Giuseppe Stragapede (e-mail: giuseppe.stragapede@uam.es).}

\vspace{-21pt}

\begin{abstract}
Analyzing keystroke dynamics (KD) for biometric verification has several advantages: it is among the most discriminative behavioral traits; keyboards are among the most common human-computer interfaces, being the primary means for users to enter textual data; its acquisition does not require additional hardware, and its processing is relatively lightweight; and it allows for \textit{transparently} recognizing subjects. However, the heterogeneity of experimental protocols and metrics, and the limited size of the databases adopted in the literature impede direct comparisons between different systems, thus representing an obstacle in the advancement of keystroke biometrics. 
% Moreover, the security challenges that keystroke biometrics promises to solve are constantly evolving and getting more sophisticated: identity fraud, account takeover, sending unauthorized emails, etc. These challenges are magnified in real applications dealing with big data (up to millions of subjects). \\
To alleviate this aspect, we present a new experimental framework to benchmark KD-based biometric verification performance and fairness based on \textit{tweet}-long sequences of variable transcript text from over 185,000 subjects, acquired through desktop and mobile keyboards, extracted from the Aalto Keystroke Databases. The framework runs on CodaLab in the form of the Keystroke Verification Challenge (KVC)\footnote{\href{https://sites.google.com/view/bida-kvc/}{https://sites.google.com/view/bida-kvc/}}. 
%By returning a series of metrics related to authentication performance and fairness, a comparative analysis of different biometric systems is made possible. %, but also a comparative assessment to position KD with respect to more exploited biometric modalities. 
Moreover, we also introduce a novel fairness metric, the Skewed Impostor Ratio (SIR), to capture \textit{inter}- and \textit{intra}-demographic group bias patterns in the verification scores.
We demonstrate the usefulness of the proposed framework by employing two state-of-the-art keystroke verification systems, \textit{TypeNet} and \textit{TypeFormer}, to compare different sets of input features, achieving a less privacy-invasive system, by discarding the analysis of text content (ASCII codes of the keys pressed) in favor of extended features in the time domain. 
Our experiments show that this approach allows to maintain satisfactory performance. 
\end{abstract}


% \begin{keywords}
% \end{keywords}

\section{Introduction}
%Keyboards are the primary means for users to enter textual data. % It is calculated that over 7 billion people are familiar with keyboards, and that at least 4 million people use it daily. It is calculated that over TB of information pass worldwide through a keyboard every day, making it a very wide and relevant, yet relatively unexplored, research area. 

\subsection{Keystroke Dynamics for Biometric Recognition}
\label{subsec:introintro}

Keystroke Dynamics (KD) refers to the typing behavior of human subjects. It is commonly regarded as a \textit{behavioral biometric} trait, similarly to voice \cite{voice}, signature \cite{TOLOSANA2022108609, 9335993}, gait \cite{PAMI_RVera2018, gait1, gait2}, touch gestures \cite{stragapede2022ijcb, stragapede2023behavepassdb}, etc. 
In comparison with its physiological counterparts such as face or fingerprint, behavioral biometrics represent a more challenging technical problem in terms of recognition performance as they are in general characterized by a higher \textit{intra-user} variability, and lower \textit{inter-user} variability. These challenges are magnified when dealing with real-life applications that have up to millions of subjects. Nevertheless, they offer the advantage of verifying identities \textit{transparently}, improving security, as well as usability \cite{ISO1998}, since they spare users from having to actively carry out a specific verification procedure (such as having their fingerprint scanned, or typing a password). 

In particular, the deployment of keystroke dynamics verification systems is also economic, as there is no need for additional hardware. Potential applications span from verifying the subject identity while they write an email or they take a test in online educational platforms (free-text format) \cite{hernandez2019edbb}, to identifying malicious users across multiple accounts based on their typing style (free-text format) \cite{typenet}, or as an additional biometric security layer on top of a traditional knowledge-based password (fixed-text format) \cite{kboc}, etc. These aspects have prompted several companies to develop commercial solutions to enhance the security of users through KD \cite{typenet}.

A coarse classification of KD can take place according to two criteria: \textit{(i)} the typology of acquisition device (keyboard): desktop or mobile. Due to differences in the pose or activity of typing subjects, more variability is commonly associated to mobile touchscreens in comparison with desktop keyboards; and \textit{(ii)} regarding the text format, which can be free, fixed, or transcript. In the first case, the text typed is not the same across different samples: consequently, data are much sparser, more unstructured, and they present a higher rate of typing errors, compared to the fixed-text case, which aims to represent for instance the case of an intruder typing the password of the victim. Finally, the transcript text could be defined as a hybrid format as the subjects are asked to read, memorize, and type a text that is presented to them.

In its simplest form, keystroke dynamics are captured as discrete time instants: the time instants a key is pressed and released (for instance in Unix time format), accompanied by the code (ASCII) of the key pressed. More complex features can be extracted from these raw data. In particular, the ASCII codes are useful for learning relations between time and spatial distributions over the keyboard layout. Nevertheless, although handled in compliance with sensitive data protection regulations \cite{GDPR2016a}, they inevitably reveal the content of the text, putting at risk the privacy of the subjects \cite{delgadosantos2021survey, melzi2022overview}. Other information such as the amount of pressure on the key or the size of the fingertip might be available depending on the specific hardware capabilities.

\subsection{Limitations of Existing Evaluation Methodologies}
\label{subsec:limitations}

In the field of keystroke biometrics, a typical obstacle for research advancement is represented by the heterogeneity of databases, experimental protocols, and metrics. In Table \ref{tab:old_dbs}, some of the most important public keystroke dynamics databases are reported in chronological order. Although the literature on keystroke biometrics is extensive, to the best of our knowledge, except very few cases \cite{typenet, typeformer}, previous systems have mostly been only evaluated with up to several hundred subjects not representing well the recent challenges that massive usage applications can face. 
In addition, most research works are mainly focused only on desktop and fixed-text scenarios. Therefore, keystroke dynamics can still be considered a biometric modality at the early stages, especially for mobile devices. In fact, for mechanical keyboards of desktop computers, more in‐depth evaluations have been conducted and commercial applications have been proposed \cite{Maiorana2021}.
Moreover, even if using the same databases, different systems proposed in the literature over the years have often been developed based on different subsets of users for development and evaluation, number of enrolment sessions, and metrics, hindering direct comparisons. In contrast, we propose a clearly defined experimental protocol based on same realistic use cases (Sec. \ref{sec:experimental_protocol}), that can be easily adopted by researchers and practitioners of the field using the provided comparison files (Sec. \ref{sec:resources}). 

\begin{table}[h!]
% \scriptsize
\centering
 \begin{tabular}{c|c|c|c|c} 
 \hline
  \hline
 \textbf{Database} & \textbf{Scenario}
 & \makecell{\textbf{No. of}\\\textbf{Subjects}} & \makecell{\textbf{Text}\\\textbf{Format}} & \makecell{\textbf{Strokes per}\\\textbf{Subject}} \\ 
 \hline
 \makecell{GREYC\\ (2009) \cite{greyckeytroke}}  & Desktop & 133 & Fixed & $\sim$800 \\
\makecell{CMU\\ (2009) \cite{cmu}} & Desktop & 51 & Fixed & $\sim$400\\ 
 \makecell{BiosecurID\\ (2010) \cite{biosecurid}} & Desktop & 400 & Free & $\sim$200 \\
 \makecell{RHU\\ (2014) \cite{rhu}} & Desktop & 53 & Fixed & $\sim$600 \\ 
 \makecell{Clarkson I\\ (2014) \cite{clarksonI}} & Desktop & 39 & \makecell{Fixed, free} & $\sim$20k\\

 \makecell{SUNY\\ (2016) \cite{sun2016shared}} & Desktop & 157 & \makecell{Transcript,\\ free} & \makecell{$\sim$17k} \\ 
 \makecell{Clarkson II\\ (2017) \cite{clarksonII}} & Desktop & 103 & Free & $\sim$125k \\
 \makecell{\textbf{Aalto Desktop}\\ \textbf{(2018) \cite{Dhakal2018}}} & \textbf{Desktop} & \textbf{168k} & \textbf{Transcript} & \makecell{$\sim$\textbf{750}}\\
 \makecell{\textbf{Aalto Mobile}\\ \textbf{(2019) \cite{palin2019people}}} & \textbf{Mobile} & \textbf{37k} & \textbf{Transcript} & \makecell{$\sim$\textbf{750}}\\
 \makecell{HuMIdb\\ (2020) \cite{acien2021becaptcha}} & Mobile & 600 & Fixed & \makecell{$\sim$20} \\

\hline
\hline
 \end{tabular}
 \caption{\small Some of the most important public keystroke dynamics databases in chronological order.}
 \label{tab:old_dbs}
\end{table}

\subsection{Fairness Considerations}
\label{subsec:fairness}

Moreover, in the context of decision-making, algorithms are vulnerable to biases that render their decisions ``unfair” \cite{10.1145/3457607, 2022_AI_SensitiveLoss_IS}. Consequently, in this context, \textit{fairness} is defined as the absence of any prejudice or favoritism toward an individual or group based on their inherent or acquired characteristics. In the last years, innumerable studies have highlighted the existence of biases in biometric systems with regard to categories such as age, gender, and ethnicity \cite{2021_TTS_Biases_Terhorst}, leading to worse decisions that affect specific demographic groups. In addition, these aspects are also relevant from the point of view of the privacy of users, as the existence of bias due to sensitive attributes often implies that the sensitive attributes themselves might be embedded in the learned representations \cite{morales2020sensitivenets, DELGADOSANTOS202230}. Therefore, the risk of leakage of some soft-biometric\footnote{Soft biometrics are physical or behavioral biometrics, or material accesories, associated with an individual, useful for recognizing an individual \cite{Dantcheva}. Examples include age, gender, ethnicity, etc. \cite{2018_TIFS_SoftWildAnno_Sosa}} information about the subjects should be assessed as well. In general terms, the existence of bias or privacy leakage in biometric systems presupposes specific patterns in the input data associated with different demographic groups. For instance, in face biometrics, the existence of biological differences between different genders, ages, or ethnic groups is a trivial hypothesis that does not need a formal demonstration. However, for many biometric modalities including KD, it is not straightforward to make similar assumptions. Nevertheless, for KD, several studies have evaluated the predictability of gender \cite{TSIMPERIDIS20184, 8966639}, age \cite{10.1007/978-3-319-91189-2_33, 10.1007/978-3-030-64758-2_4}, both \cite{10.1145/3099023.3099105, electronics10070835}, and even emotions \cite{doi:10.1080/0144929X.2014.907343} and mother tongue \cite{telecom4030021}. In light of this, in this article we propose an experimental framework designed to highlight potential gender and age biases in the scores, which are still mainly unexplored aspects for KD on such a large scale. Within the current work, the focus is limited to age and gender because other potential sources of bias (such as the subject mother tongue, the device used, or the degree of familiarity of the subjects with keyboards) were not reported for most subjects in the raw databases.

% Differently from face biometrics, in which the existence of gender and age differences in the input data needs not any demonstration, keystroke dynamics, as well as all traits in which 

\subsection{Contributions}
\label{subsec:contributions}

In brief, the main contributions of this article can be summarized as follows:
\begin{itemize}
    % \item We propose a novel open experimental framework to benchmark KD for biometric verification, running on CodaLab, in the form of a challenge, the Keystroke Verification Challenge. We consider several metrics that quantify the authentication performance as well as the fairness of biometric systems built on KD (Sec. \ref{sec:evaluation_description}). To create the framework, we considered two of the largest public databases of keystroke dynamics up to date, the Aalto Desktop \cite{Dhakal2018} and Mobile \cite{palin2019people} Keystroke Databases. 
    \item We propose a novel experimental framework to benchmark KD for biometric verification, which, to the best of our knowledge, is still lacking in this field. The framework is provided in the form of the Keystroke Verification Challenge (KVC)\footnote{The challenge is held within the \href{http://bigdataieee.org/BigData2023/}{2023 IEEE International Conference on Big Data (IEEE Big Data)}, Sorrento, Italy, December 15$^{th}$-18$^{th}$, 2023. After such term, the challenge will be made ongoing so that the proposed framework can become a useful resource for all researchers and practitioners of the field. Website: \href{https://sites.google.com/view/bida-kvc/}{https://sites.google.com/view/bida-kvc/}}, hosted on CodaLab\footnote{\href{https://codalab.lisn.upsaclay.fr/competitions/14063/}{https://codalab.lisn.upsaclay.fr/competitions/14063/}}. The CodaLab platform returns several metrics (Sec. \ref{sec:evaluation_description}) that quantify the recognition performance as well as the fairness of biometric systems. To create the framework, we consider two of the largest public databases of keystroke dynamics up to date, the Aalto Desktop \cite{Dhakal2018} and Mobile \cite{palin2019people} Keystroke Databases, extracting datasets that guarantee a minimum amount of data per subject, age and gender annotations, absence of corrupted data, and that avoid too unbalanced subject distributions with respect to the considered demographic attributes.
    \item We illustrate the main aspects of the proposed framework by considering two recent state-of-the-art keystroke biometric systems, TypeNet \cite{typenet}, and TypeFormer \cite{typeformer, 10042710}. To this end, we propose a thorough analysis considering four different sets of features (Sec. \ref{subsec:evaluation}) towards more privacy-preserving biometric systems not requiring the ASCII code, which would reveal the text content, as an input feature.   
    Our experiments show that by removing spatial information of the key location on the keyboard layout (ASCII code) in favor of additional features in the time domain, an acceptable level of performance is maintained.
    \item A comparative analysis of keystroke dynamics verification systems in desktop and mobile scenarios is provided (Sec. \ref{subsec:biometric_verification}.
    %that can keep the recognition performance but keeping the privacy of the subjects as it does not include the typed message (the ASCII code)
%We consider different sets of features towards a more privacy-preserving keystroke verification system, that does not require to acquire the ASCII code, which would reveal the text content, as an input feature.
    \item We propose a new metric, the Skewed Impostor Ratio (SIR), useful to quantify how harder is for the classifier a pairwise comparison between subjects belonging to the same demographic group in relation with comparisons of subject belonging to different groups.
\end{itemize}
% We propose a new metric, the Skewed Impostor Rate (SIR), useful to quantify how much harder is for the classifier a pairwise comparison between subjects belonging to the same demographic group in relation with comparisons of subject belonging to different groups.

The remainder of the article is organized as follows: first, the resources provided within the proposed experimental framework are described (Sec. \ref{sec:resources}). Then, Sec. \ref{sec:evaluation_description} includes a detailed presentation of the evaluation protocol of the experimental framework and challenge, whereas Sec. \ref{sec:metrics} presents the metrics adopted, including the definition of SIR, a novel metric proposed in this article. Sec. \ref{sec:biometric_verification_systems} provides an overview of the two biometric systems, TypeNet \cite{typenet} and TypeFormer \cite{typeformer}, utilized to validate the framework, followed by Sec. \ref{sec:experimental_protocol}, in which the set of experiments for privacy-enhancement is illustrated. Finally, Sec. \ref{sec:experimental_results} and Sec. \ref{sec:conclusions} respectively contain the analysis of the results obtained and the article conclusive remarks.

\section{Resources Provided}
\label{sec:resources}

The proposed experimental framework is based on the two most complete and large-scale public databases of free-text keystroke dynamics up to date, collected by the User Interfaces\footnote{\href{https://userinterfaces.aalto.fi/}{https://userinterfaces.aalto.fi/}} group of the Aalto University (Finland). The two databases are collected respectively in a desktop\footnote{\href{https://userinterfaces.aalto.fi/136Mkeystrokes/}{https://userinterfaces.aalto.fi/136Mkeystrokes/}} \cite{Dhakal2018} and mobile\footnote{\href{https://userinterfaces.aalto.fi/typing37k/}{https://userinterfaces.aalto.fi/typing37k/}} \cite{palin2019people} acquisition environment, including respectively around 168,000 and 60,000 subjects, thus representing well the typical challenges related to massive application usage. Each of the acquisition sessions contains a sentence of transcript text (variable content, but not fully free-text). The data were captured through a web application in an unsupervised way under realistic scenarios. Subjects were asked to read, memorize, and type in their device English sentences that were randomly selected from a set of 1,525 sentences. Subject metadata such as age and gender are self-reported during the data acquisition.

The two databases have been processed to arrange the data in a convenient format for the analysis of KD. 
The raw data acquired consist of the timestamp of the instant a key is pressed, the timestamp of the instant the key is released, and the key ASCII code. After discarding some of the subject data due to insufficient acquisition sessions per subject (less than 15 per subject), the two databases as downloaded have been rearranged to form four datasets:
\begin{itemize}
    \item Desktop Dataset:
    \begin{enumerate}
        \item Development set: 115,120 subjects provided in a single .npy file that contains a Python nested dictionary (subject IDs: session IDs: data). Average session length: 48.65 ($\sigma$ = 18.50) characters typed.
        \item Evaluation set: data from 15,000 subjects, provided in a single .npy file that contains a shallow Python dictionary (sessions IDs: data). Average session length: 48.77 ($\sigma$ = 18.64) characters typed.
    \end{enumerate}
    \item Mobile Dataset:
    \begin{enumerate}
        \item Development set: 40,639 subjects provided in a single .npy file that contains a Python nested dictionary (subject IDs: session IDs: data). Average session length: 48.59 ($\sigma$ = 21.84) characters typed.
        \item Evaluation set: data from 5,000 subjects, provided in a single .npy file that contains a shallow Python dictionary (sessions IDs: data). Average session length: 47.98 ($\sigma$ = 20.93) characters typed.
    \end{enumerate}
\end{itemize}

The proposed experimental framework follows an open-set learning protocol, in other words, the subjects in the development and evaluation sets are different\footnote{As the datasets are fixed with a single choice of subjects, the conclusions obtained from the reported results can be impacted. To verify that the score fluctuations are not significant, we randomly split the evaluation score lists (Sec. }\ref{sec:evaluation_description}) in 10 subsets. Then, we computed the global EER (Sec. \ref{sec:metrics}) for each of the random subsets, to provide mean and standard deviation. As an example, we report the following values for TypeFormer \textit{5F} (Sec. \ref{subsec:evaluation}): $\mu = 12.949\%$, $\sigma = 0.090\%$for desktop, $\mu = 10.164\%$, $\sigma = 0.073\%$ for mobile. Please refer to Sec. \ref{sec:metrics} and \ref{sec:experimental_results} for details about the metrics and results. A validation set is not explicitly provided, but it can be obtained from the development set according to different training approaches. 

% \subsection{Demographic distribution of datasets}
Table \ref{table:demo} shows the demographic distribution of the datasets provided in the KVC. The subjects have been divided into six age groups (10 - 13, 14 - 17, 18 - 26, 27 - 35, 36 - 44, 45 - 79). The evaluation sets are balanced with respect to gender. The gender and age labels of the development set are released together with the data. %This information will not be provided to the participants, but it will be used for further analysis in the competition summary paper.

% Table \ref{table:demo} shows the demographic distribution of the datasets provided in the competition. The subjects have been divided into six age groups (10 - 13, 14 - 17, 18 - 26, 27 - 35, 36 - 44, 45 - 79). The evaluation sets are balanced with respect to gender. Table \ref{table:demo} shows the demographic distribution of the datasets provided in the competition. The subjects have been divided into six age groups (10 - 13, 14 - 17, 18 - 26, 27 - 35, 36 - 44, 45 - 79). The evaluation sets are balanced with respect to gender. 

\begin{table}[!t]
\centering
\caption{\small Demographic distributions of the provided datasets. The rows represent different age groups, while the columns represent genders. The evaluation sets are balanced with respect to gender.}
\textbf{Task 1: Desktop Dataset}\\~\\
\begin{minipage}{.5\linewidth}
    \centering
    \textbf{Development Set}\\
    \label{tab:first_table}
\begin{tabular}{c|cc} 
\toprule
 & \textbf{Male}  &  \textbf{Female} \\   
\hline
 \textbf{10 - 13} & 4,336 & 5,420 \Tstrut\\ 
 \textbf{14 - 17} & 10,993 & 8,336 \Tstrut\\ 
 \textbf{18 - 26} & 25,752 & 24,315 \Tstrut\\ 
 \textbf{27 - 35} & 9,607 & 12,281 \Tstrut\\  
 \textbf{36 - 44} & 2,143 & 5,331 \Tstrut\\ 
 \textbf{45 - 79} & 1,182 & 5,424 \Tstrut\\ 
\bottomrule
\end{tabular}\\
\vspace{0.1cm}
Total Labelled: 115,120,\\ Total unlabeled: 0\\
\end{minipage}\hfill
\begin{minipage}{.5\linewidth}
    \centering
     \textbf{Evaluation Set}\\
     \vspace{0.05cm}

    \label{tab:second_table}
\begin{tabular}{ c|c c } 
\toprule
 & \textbf{Male}  &  \textbf{Female} \\   
\hline
 \textbf{10 - 13} & 1,085 & 1,085 \Tstrut\\ 
 \textbf{14 - 17} & 1,861 & 1,861 \Tstrut\\ 
 \textbf{18 - 26} & 1,861 & 1,861 \Tstrut\\ 
 \textbf{27 - 35} & 1,861 & 1,861 \Tstrut\\  
\textbf{36 - 44} & 536 & 536 \Tstrut\\ 
 \textbf{45 - 79} & 296 & 296 \Tstrut\\ 
\bottomrule
\end{tabular}\\
\vspace{0.1cm}
Total Labelled: 15,000,\\ Total unlabeled: 0\\
\end{minipage}
~\\~\\
 \textbf{Task 2: Mobile Dataset}\\     
~\\
\begin{minipage}{.5\linewidth}
    \centering
\textbf{Development Set}\\
\begin{tabular}{ c|c c } 
\toprule
 & \textbf{Male}  &  \textbf{Female} \\   
\hline
 \textbf{10 - 13} & 622 & 800 \Tstrut\\ 
 \textbf{14 - 17} & 1,537 & 1,516 \Tstrut\\ 
 \textbf{18 - 26} & 4,359 & 8,999 \Tstrut\\ 
 \textbf{27 - 35} & 1,343 & 4,002 \Tstrut\\  
\textbf{36 - 44} & 382 & 1,333 \Tstrut\\ 
 \textbf{45 - 79} & 200 & 739 \Tstrut\\ 
 \bottomrule
\end{tabular}\\
\vspace{0.1cm}
 Total Labelled: 25,832,\\ Total unlabeled: 14,807\tablefootnote{Although unlabeled, we opted to include these subjects to maximize the size of the provided dataset.}\\
\end{minipage}\hfill
\begin{minipage}{.5\linewidth}
\centering
 \textbf{Evaluation Set}\\
\vspace{0.05cm}
\begin{tabular}{ c|c c } 
\toprule
 & \textbf{Male}  &  \textbf{Female} \\   
\hline
 \textbf{10 - 13} & 254 & 254 \Tstrut\\ 
 \textbf{14 - 17} & 618 & 618 \Tstrut\\ 
 \textbf{18 - 26} & 843 & 843 \Tstrut\\ 
 \textbf{27 - 35} & 547 & 547 \Tstrut\\  
\textbf{36 - 44} & 156 & 156 \Tstrut\\ 
 \textbf{45 - 79} & 82 & 82 \Tstrut\\ 
\bottomrule
\end{tabular}\\
\vspace{0.1cm}
Total Labelled: 5,000,\\ Total unlabeled: 0\\
\end{minipage}

\label{table:demo}
\end{table}

The evaluation sets are separated by scenario (desktop and mobile), and they are provided in the form of two shallow Python dictionaries containing independent sessions. Such data are accompanied by the respective lists of pairwise comparisons to be carried out. Two Python script files are provided to load the data, and run the comparisons, generating a text file with the scores of each comparison, ready to be submitted to CodaLab for scoring. To push forward the state of the art and deepen the knowledge on the topic, the proposed protocol is designed for researchers working on KD as a novel tool to evaluate different approaches (pre-processing of input features, classifier architectures, learning approaches, etc.) for different goals (biometric recognition and fairness improvement) under the same experimental conditions, considering various metrics.

% \subsection{Start-off Scripts}
% Two scripts are provided with the data to simplify the participation:
% \begin{itemize}
%     \item A training script that contains a very simple \texttt{Pytorch} implementation of a recurrent neural network (RNN) followed by a linear layer. The model is trained with a contrastive loss function. 20\% of the subjects in the development set (selected randomly) are used for validating the model. At the end of each epoch, the model is saved (and overwritten) if a lower value of EER on the validation set is achieved. The raw data are processed to extract the duration of each press and the ASCII code, and then zero-padded or sliced to a fixed sequence length of 100. For each input sequence, the output of the model is a 32-value array (embedding). The goal is mapping embeddings belonging to the same subject close to each other, while distancing embeddings belonging to different subjects. The training script works for both development sets (desktop and mobile). 
%     \item An evaluation script that loads the trained model, extracts the features described above, computes the embeddings in the specific list of comparisons (desktop or mobile, according to the scenario variable), and it computes the scores for each of the comparisons. Then, a text file containing the comparisons is saved, ready to be submitted to CodaLab for evaluation.
% \end{itemize}

\section{Evaluation Description}
\label{sec:evaluation_description}
The design and the implementation of the evaluation protocol described in this section represents a significant novelty aspect proposed in the current work.

The two tasks (desktop and mobile) are structured similarly, and they are designed for a biometric verification protocol. In other words, a score between 0 and 1 related to a single comparison of two biometric samples will be produced (1: same identity, 0: different identities). It is a binary classification problem, as it is not necessary to ascertain to which identity a specific biometric sample belongs to (identification). In this experimental framework, a biometric sample corresponds to an acquisition session. 

The total number of 1 vs 1 session-level comparisons is as follows:
\begin{itemize}
    \item Task 1 (Desktop): 2,250,000 comparisons, involving 15,000 subjects not included in the development set.
    \item Task 2 (Mobile): 750,000 comparisons, involving 5,000 subjects not included in the development set.
\end{itemize}

\begin{figure}[t]
\centering
  \centering
  \includegraphics[width=\linewidth]{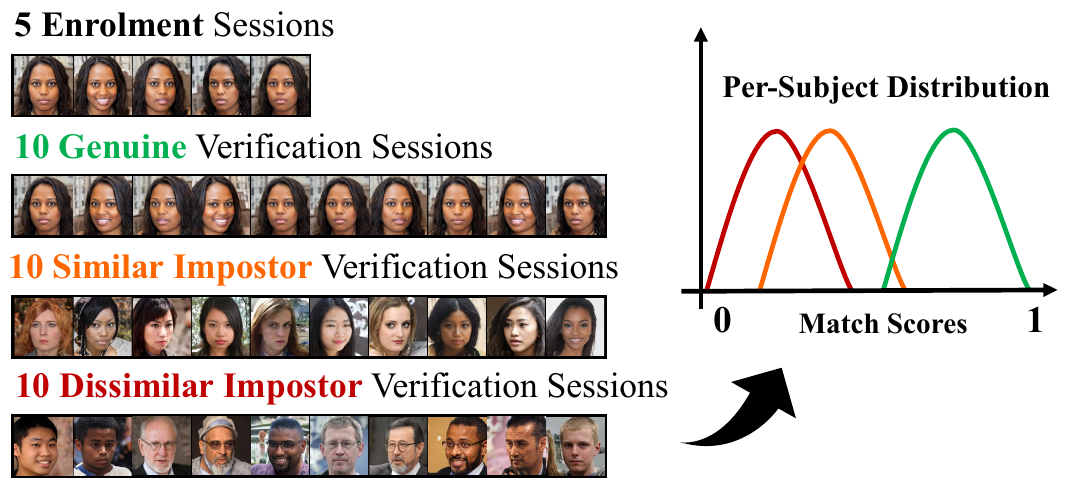}
\caption{\small Each one of the verification sessions is compared with each of the enrolment sessions. For an easier comprehension, examples of faces showing gender and age are included instead of keystroke examples. Then, the scores generated are averaged over the enrolment session, leading to three distributions: genuine (green), similar impostor (orange), and dissimilar impostors (red).}
\label{fig:distr}
\end{figure}

The design of the comparisons is illustrated in Fig. \ref{fig:distr}. For each subject, there are 5 enrolment sessions and 10 verification sessions, leading to 50 1vs1 comparisons, which are averaged over the 5 enrolment sessions generating 10 genuine scores per subject. In a similar manner, 20 impostor scores per subject are generated. The impostor sessions are divided into two groups: 10 \textit{similar} impostor scores, for which the verification sessions are randomly selected from subjects belonging to the same demographic group (same gender and age); 10 \textit{dissimilar} impostor scores, in which the verification sessions are all randomly selected from subjects of different gender and age intervals \footnote{To verify that the score fluctuations due to randomness are not significant, we repeated the choice 10 times with 10 different seeds. As an example, we report the following evaluation values for TypeFormer \textit{5F} (Sec. \ref{subsec:evaluation}) in terms of global EER (Sec. }\ref{sec:metrics}): $\mu = 12.410\%, \sigma = 0.036\%$ in the desktop case, $\mu = 9.444\%, \sigma = 0.045\%$ in the mobile case. Please refer to Sec. \ref{sec:metrics} and \ref{sec:experimental_results} for details about the metrics and results.

Based on the described evaluation design, following \cite{typenet, typeformer}, we consider two cases for evaluating the system:
\begin{itemize}
    \item Global distributions: this case corresponds to dividing all scores into two groups, genuine and impostor scores, regardless of which subject they belong too. This case corresponds to a having a fixed, pre-determined threshold, implying a simpler deployment of the biometric system. In order to assess the performance of the biometric system, this choice means setting one single threshold for all comparisons to obtain a decision. 
    \item Mean per-subject distributions: the optimal threshold is computed at subject-level, considering the 30 verification scores as described above. This choice corresponds to
    providing the system with more flexibility, so that it can adapt to user-specific distributions \cite{Fierrez-Aguilar2005_ScoreNormalization, Fierrez-Aguilar2005_AdaptedMultimodal}. In a real-life use case, this would require processing the subject's enrolment samples to establish a threshold, and it can be done as follows: acquiring various enrollment samples, from which to derive a genuine subject-specific score distribution by considering pairwise comparisons between enrollment samples; considering a pool of samples from different subjects, from which to derive an impostor subject-specific score distribution by considering pairwise comparisons with the genuine enrollment samples; computing a subject-specific threshold based on the two distributions. It is important to highlight that this does not require re-training or fine-tuning the biometric system using subject-specific data. Then, all metrics computed per-subject are averaged considering all subjects in the evaluations set to obtain the values displayed. Generally, the verification performance of the system benefits from considering a different threshold per user. 
\end{itemize}

\section{Metrics Adopted}
\label{sec:metrics}
% \par Keystroke verification systems have evolved (and keep evolving) together with the available technology (deep learning-based classifiers), improving the recognition performance, and a typical issue of the field of keystroke biometrics is the heterogeneity of databases, experimental protocols, and metrics. Therefore, a rigorous comparison between the different performance values is a difficult operation. With this challenge we aim to provide a complete panorama of the state of the art in the field of free-text keystroke dynamics. 
%In addition, we have recently developed a Transformer model, named TypeFormer, for the purpose of keystroke-based subject authentication \cite{typeformer}. Consequently, we aim to compare the performance of our system with different approaches and advance the state of the art. 
%The performance of the systems proposed by the participants will be evaluated considering several metrics, popular in the field of biometrics. 
% Table \ref{tab:all_metrics} shows the metrics presented in Sec. \ref{sec:metrics}. The trends highlighted by the graphs are consistent throughout different metrics. 

Within the years, several metrics have been proposed for biometric verification. The common aspect of all metrics is that they are based on the (normalized) scores that are typically generated by pairwise comparisons of biometric data. However, a comparison between systems is often a difficult operation if they are evaluated according to different metrics. Moreover, the attention of the scientific community has recently shifted towards the evaluation of the \textit{fairness} of systems \cite{9975333}. Consequently, based on the scores, we also provide an initial assessment of this important aspect which, to the best of our knowledge, is still an unexplored aspect of KD. An overview of all metrics considered is provided below. The scores will be computed considering global and mean per-subject distributions. All the presented metrics are returned by the KVC CodaLab scoring program, easily allowing experimental analyses of multiple aspects, such as different sets of input features for privacy-enhancement (Sec. \ref{sec:experimental_results}). To the best of our knowledge, these scenarios have not been proposed in previously existing literature.

\subsection{Verification metrics}
In biometrics, a false match (FM) is defined as a comparison decision of a match for a biometric probe and a biometric reference that are from different biometric capture subjects, while a false non-match (FNM) is defined as comparison decision of non-match for a biometric probe and a biometric reference that are from the same biometric capture subject and of the same biometric characteristic. The rate respectively associated with FMR (FNMR) corresponds to proportion of the completed biometric non-mated (mated) comparison trials that result in a false match (non-match) \cite{ISObiometrics}.

%In biometrics, a false rejection (FR) is defined as a genuine subject recognized as an impostor, while false acceptance (FA) corresponds to an impostor subject recognized as genuine. A true rejection (TR) and a true acceptance (TA) are respectively an impostor and a genuine subject classified correctly. The rates respectively associated with each one of those quantities (FRR, FAR, TRR, TAR) are computed over all access attempts. 

\subsubsection*{Equal Error Rate (EER)} The EER describes the point in which the FMR and FNMR curves intersect. The two rates typically have opposite trends with respect to the threshold setting (in the case of genuine scores closer to 1, and impostor scores closer to 0, as the threshold of a biometric system increases, the FMR will drop and the FNMR curve will rise). On the DET curve (Sec. \ref{subsec:curves}), which is the plot of FNMR against FMR, at various threshold settings, it corresponds to the point where $y = x$.

\subsubsection*{False Non-Match Rate at X\% False Match Rate (FNMR @ X\% FMR)} We consider $X = 1\%, 10\%$. This also corresponds to a point on the DET curve. This metric expresses a trade-off between security and usability \cite{ISO1998}. In fact, while from the point of view of security the priority is avoiding intrusions, denying the access to the genuine subject a large number of times would generate frustration and highly impacts the usability of the system. In this case, the threshold is set to X = 1\%, 10\% of FMR (rejection of 99\%, 90\% of impostor attempts, respectively), aiming to minimize the FNMR.

\subsubsection*{Area Under the Receiver Operating Characteristic (ROC) Curve (AUC)} The ROC curve (Sec. \ref{subsec:curves}) is the plot of the TMR (True Match Rate) against FMR, at various threshold settings. A true match corresponds to the case of a genuine subject recognized as such. By definition, the TMR and the FNMR sum to 1. A perfect classifier has an Area Under the ROC Curve (AUC) of 1. 

\subsubsection*{Accuracy} The accuracy is computed as the fraction of correctly classified attempts at a given discrimination threshold $\tau$, corresponding, in the current work, to the EER threshold.

\subsubsection*{Rank-\textit{n}} This metric concerns the identification of subjects (i.e., 1 to many comparisons), therefore assessing a different scenario from the previous metrics, which refer to the case of verification (i.e., binary classification). Starting from the comparison of biometric enrolment samples with \textit{N} biometric samples including a genuine one, it represents the rate to which the genuine scores fall within the best \textit{n} matches. In the proposed framework, this metric is computed by considering separately each one of the genuine verification sessions against all 20 impostor sessions available for each subject. The returned rank-\textit{n} values are averaged over the 10 genuine verification sessions (\textit{n} = 1).

\subsection{Curves}
\label{subsec:curves}
\subsubsection*{Score Histograms}
They are computed considering the global genuine and impostor distributions. It is necessary to have a clear separation between the two, with the genuine distribution shifted toward 1, and the impostor one toward 0. A small overlap of the tails corresponds to a better performance of the system.

% \subsubsection*{EER Curves} The EER is defined as the point where the FRR and the FAR curves intersect. The curves are obtained by plotting all possible FRRs and FARs at different thershold settings. The curves have opposite trends, with the FRR increasing, and FAR decreasing.

\subsubsection*{Detection Error Trade-off (DET) curve} It is the plot of FMR against FNMR, at various threshold settings, typically on a non-linear scale. As the threshold decreases, the amount of false matches (impostor subjects classified as genuine) increases, and the number of false non-matches decreases (genuine subjects classified as impostor). The closest the DET curve to the bottom left corner, the better the biometric system will be.

\subsubsection*{ROC Curve} It is the plot of the TMR against FMR, at various threshold settings. 

\subsection{Fairness metrics}
\subsubsection*{Standard Deviation (STD) of EER by demographic group} It considers the demographic differential assessment by calculating the standard deviation in accuracy performance between all demographic groups at a given discrimination threshold $\tau$ (in this work corresponding to the global EER threshold). The STD expresses a measure of how dispersed the values is in relation to the mean. The optimal STD value is 0\%.
\subsubsection*{Skewed Error Ratio (SER) of EER by demographic group} Skewness is a measure of the asymmetry of a distribution. Similarly to the STD, SER is computed across demographic subsets as the ratio between the greatest and smallest error scores. It mainly represents the difference between the sensitive attribute with the best and worst performance. The larger the value, the greater the difference in the algorithm’s discrimination towards a certain attribute. The optimal SER value is 1.

\subsubsection*{Fairness Discrepancy Rate (FDR)} It was proposed in \cite{9507539}. It considers the FMR and FNMR trade-off in the demographic differential assessment by calculating the max difference in TMR and TNMR performance between any two demographic groups $d_{i}$ and $d_{j}$ and a given discrimination threshold $\tau$. In the current work, $\tau$ corresponds to the point of FMR = 1\%. Those differences are then weighted by parameters $\alpha$ and $\beta = 1-\alpha$, which represent the level of concern applied to differences in FMR and FNMR respectively. It is advantageous due to its formulation, that encompasses two different metrics, and its flexibility given by the parameters $\alpha$ and $\beta$. It can be plotted as a function of every possible threshold value. Consequently, it is necessary to have access to the ground truth. The FDR ranges from 0 (fair) to 1 (unfair).

\subsubsection*{Inequity Rate (IR)} It was proposed in \cite{grother2019face}. It is computed considering the ratio differences between minimum and maximum FMR and FNMR per demographic groups $d_{i}$ and $d_{j}$ for $\tau$ corresponding to the point of FMR = 1\%. Although spanning different ranges from, its characteristics are similar to those of FDR. The IR ranges from 0 (fair) to infinite (unfair).

% \begin{equation*}
%     A(\tau) = \frac{max_{d_{i}} FMR_{d_{i}(\tau)}}{min_{d_{i}} FMR_{d_{i}(\tau)}} 
% \end{equation*}
% \begin{equation*}
%     B(\tau) = \frac{max_{d_{i}} FNMR_{d_{i}(\tau)}}{min_{d_{i}} FNMR_{d_{i}(\tau)}} 
% \end{equation*}
% \begin{equation*}
%     IR = A(\tau)^{\alpha} B(\tau)^{1-\alpha}  
% \end{equation*}

\subsubsection*{Gini Aggregation Rate for Biometric Equitability (GARBE)} It was proposed in \cite{howard2022evaluating} to overcome the limitations of FDR, and IR. In fact, the former does not scale the values of FMR and FNMR to the same order of magnitude, whereas the latter has no theoretical upper bound and may have a denominator equal to zero. GARBE is inspired in the mathematics of the Gini coefficient, computed for $x = \{\textrm{FMR}, \textrm{FNMR}\}$ as:
\begin{equation*}
    G_{x}(\tau) = \left( \frac{n}{n-1} \right) \left( \frac{\sum_{i=1}^{n}\sum_{j=1}^{n} |r_{i}-r_{j}|}{2 n^{2}\bar{r}} \right) 
\end{equation*}
where $r_{i}, r_{j} \in \{\textrm{FMR}_{d_{i}} @\tau|d_{i} \in \mathbb{D} \}$ for $r = \textrm{FMR}$, $r_{i}, r_{j} \in \{FNMR_{d_{i}} @\tau|d_{i} \in \mathbb{D} \}$ for $r = \textrm{FNMR}$, $\tau$ is the decision threshold (FMR = 1\%), $d_i$ refers to a demographic group from the set $\mathbb{D}$ of demographic groups, and $n$ is the number of demographic groups. These Gini coefficients are combined as follows:
\begin{equation*}
    \textrm{GARBE}(t_{z}) = \alpha G_{\textrm{FMR}}(t_{z}) + (1-\alpha) G_{\textit{FNMR}}(t_{z})
\end{equation*}
where $\alpha$ represents the level of concern applied to differences in FMR and FNMR respectively. The GARBE ranges from 0 (fair) to 1 (unfair).

\subsubsection*{Skewed Impostor Ratio (SIR)} This is a novel metric proposed in this article. Normalized impostor scores are grouped according to a specific attribute (age or gender). For instance, considering age, it is possible to group the comparisons as follows: `10-13 vs 10-13', `10-13 vs 14-17', `10-13 vs 18-26', and so on, considering all combinations. For each score group combination, the average value is taken. Then, all values can be arranged in a matrix, which is symmetric. The elements on the main diagonal represent comparisons between the different subjects belonging to the the same age or gender group (`10-13 vs 10-13', `14-17 vs 14-17', etc.), while all other elements are obtained from the remaining cross-group comparisons. The ratio between the mean value of the elements in the main diagonal, and the remaining non-duplicated elements, is finally computed as a percentage. Such value expresses how harder is a comparison between different subjects belonging to the same demographic group in comparison to subjects belonging to different ones, quantifying to which extent demographic information is retained in the scores. It can be formulated as follows:
\begin{equation*}
\textrm{SIR} = 100 \left(\frac{\mu (s_{ii})}{\mu (s_{ij, i\neq j})}-1\right), i,j = 1, ..., n
\end{equation*}
where \textit{n} is the number of demographic groups (\textit{n} = 6 for age, \textit{n} = 2 for gender), and \textit{s} is the average score for a specific type of comparisons. The optimal SIR value is 0 (\%), in case of no skew between impostor scores regardless of their demographic group.
In comparison with the previous metrics, the advantages are as follows:
\begin{itemize}
    \item It is not necessary to select a threshold value.
    \item It focuses on both \textit{intra-group} and \textit{inter-group} relations, highlighting the differences in the two cases. If the differences between the two cases are not significant nor consistent, then the system is bias-free.
    \item It is not necessary to have access to the system, which can be treated as a \textit{black box}. It is sufficient to run a significant number of appropriately distributed comparisons.
    \item By considering the entire matrix as described above, it is possible to focus on comparisons between specific groups, gaining some precious insights about the system and the similarities between demographic groups.
\end{itemize}

\section{Biometric Verification Systems}
\label{sec:biometric_verification_systems}
Throughout the proposed framework, we evaluate two recent state-of-the-art deep-learning models:
\begin{itemize}
    \item TypeNet (2021) \cite{typenet}: a Long-Short Term Memory (LSTM) Recurrent Neural Network (RNN), trained with triplet loss. In this case, we consider input sequences of 150 characters typed. TypeNet is implemented in \texttt{Tensorflow} \cite{tensorflow2015-whitepaper}.
    \item TypeFormer (2023) \cite{typeformer, 10042710}: a novel transformer architecture consisting in a temporal and a channel module enclosing two LSTM RNN layers, a Gaussian Range Encoding (GRE), a multi-head self-attention mechanism, and a block-recurrent transformer structure. TypeFormer is also trained with triplet loss. In this case, we consider input sequences of 50 characters typed. TypeFormer is implemented in \texttt{PyTorch} \cite{pytorch}.
\end{itemize}
Both approaches utilize Distance Metric Learning (DML) \cite{2022_PR_SetMargin_Morales}. The fundamental concept of DML involves training a model that transforms input data into a new feature space, enabling straightforward distances to be used for analyzing and leveraging the ``semantic'' arrangement of the input space \cite{hadsell2006dimensionality}. A DML approach aims to establish a neighborhood structure in the feature space by considering the relationship between intra-class (distances among samples from the same class) and inter-class (distances among samples from different classes) distances. In an ideal feature space, samples from the same class will remain in close proximity, while samples from different classes will be distinctly separated. Following this idea, the input sequences obtained from all sessions are transformed into feature embeddings, that are expected to have lower Euclidean distances if belonging to the same subject, higher otherwise. In the test stage, the distances obtained from the comparisons of feature embeddings corresponding to each of the test sessions are normalized, and then they are subtracted from 1 in order to transform them into similarity scores.

\section{Experimental Protocol}
\label{sec:experimental_protocol}
\subsection{Evaluation of Privacy-Enhancing Input Features}
\label{subsec:evaluation}

\begin{figure}
\centering
  \centering
  \includegraphics[width=\linewidth]{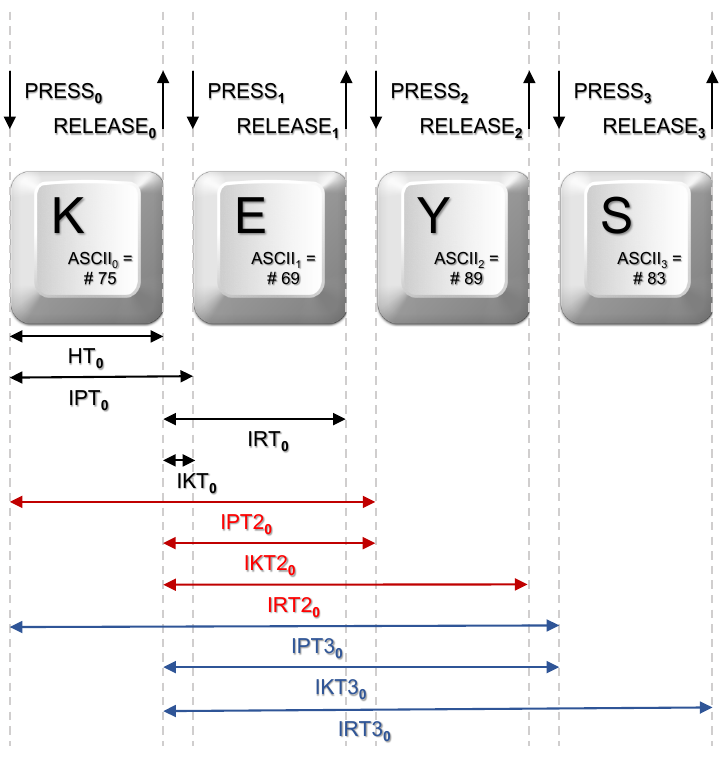}
\caption{\footnotesize A diagram representing the initial feature extraction process for the time instant $t = 0$. HT = Hold Time; IPT = Inter-Press Time; IRT = Inter-Release Time; IKT  = Inter-Key Time; IPT2 = Inter-Press Time with second following key; IRT2 = Inter-Release Time with second following key; IKT2  = Inter-Key Time with second following key; IPT3 = Inter-Press Time with third following key; IRT3 = Inter-Release Time with third following key; IKT3  = Inter-Key Time with third following key.}
\label{fig:features}
\end{figure}

The two biometric systems considered, TypeNet and TypeFormer, take as input the same set of features extracted from the raw data (Unix timestamps of the actions of pressing and releasing a key), which include:
\begin{itemize}
    \item[\textit{(i)}] Hold Time (HT): time interval between the release and press instants of a given key, expressed in seconds.
    \item[\textit{(ii)}] Inter-Press Time (IPT): time interval between two consecutive press actions, expressed in seconds.
    \item[\textit{(ii)}] Inter-Release Time (IRT): time interval between two consecutive release actions, expressed in seconds.
    \item[\textit{(iv)}] Inter-Key Time (IKT): time interval between a release and the following press action, expressed in seconds.
    \item[\textit{(v)}] ASCII code (ASCII), normalized by dividing it by 255.
\end{itemize}
Such features are graphically represented in black in Fig. \ref{fig:features}. Input features \textit{(i)}-\textit{(iv)} are useful to capture the typing behavior of the user in the time-domain, whereas the ASCII code \textit{(v)} describes the spatial relations due to the location of the key pressed on the keyboard layout. However, although handled in compliance with sensitive data protection regulations \cite{GDPR2016a}, the acquisition and processing of the ASCII codes inevitably reveals the content of the text, putting at risk the privacy of the users. Consequently, in this experiment, we strive to remove the ASCII code information in order to make the system content-agnostic, and consequently more privacy preserving. The sets of experiments run can be summarized as follows:
\begin{itemize}
    \item[1.] First, to evaluate and compare TypeNet and TypeFormer, we consider their original set of features (the experiment is named \textit{5F}, where ``\textit{F}'' stands for ``features''), marked in black in Fig. \ref{fig:features}.
    \item[2.] Then to quantify the importance of the ASCII code information, we remove the ASCII code information from the original set of features (experiment \textit{4F}).
    \item[3.] We consider an extended set of time-domain features. Several studies have in fact shown the usefulness of considering groups of keys typed such as digraphs, trigraphs, and \textit{n}-graphs \cite{9853506}. By considering not only adjacent keys, but groups of three keys (Fig. \ref{fig:features}, in red and blue), we obtain a set of 10 features (experiment \textit{10F}): 
    \begin{equation*}
        [\textrm{HT}, \textrm{IPT}, \textrm{IRT}, \textrm{IKT}, \textrm{IPT2}, \textrm{IRT2}, \textrm{IKT2},
    \end{equation*}
    \begin{equation*}
        \textrm{IPT3}, \textrm{IRT3}, \textrm{IKT3}]
    \end{equation*}

    \item[4.] We consider the extended set of time-domain features, together with the ASCII code (experiment \textit{11F}).
\end{itemize}
In each case, the deep learning models are trained from scratch.
% 
% We designed four experiments per model in order to assess the performance of different set of features: \textit{(i)} original set of features (\textit{5F}), original set of features without ASCII (\textit{4F}), proposed set of features (\textit{10F}), proposed set of features plus ASCII (\textit{11F}).  

\subsection{Model Training}
The training of both models takes place on the KVC development set considering identical settings to those described in their respective papers. The only differences are related to the division into training and validation sets. For TypeNet, we consider a subset of 400 subjects to validate the model at the end of each training epoch in terms of average EER per subject. This choice is justified by the experimental protocol followed in \cite{typenet}. For TypeFormer, we consider an 80\%-20\% train-validation division of the KVC development set, and we adopt the global EER as validation metric. According to these validation metrics, the best-performing epoch model is saved in each case.

\section{Experimental Results}
\label{sec:experimental_results}

\subsection{Biometric Verification}
\label{subsec:biometric_verification}
The results of the experiments are reported in Table \ref{tab:all_metrics}. The table is divided into two parts, each one corresponding to one scenario: desktop and mobile. Each half can be further divided into the two cases considered: results obtained in the global genuine and impostor distributions (see Sec. \ref{sec:evaluation_description}), and results considering the mean values obtained for per-subject genuine and impostor distributions. Each row shows a different system, TypeNet or TypeFormer, trained on a different set of input features, according to the 4 experiments described in Sec. \ref{subsec:evaluation}, while the different metrics are reported along the columns.

\begin{table*}[hbt!]
\centering
\scriptsize
\caption{\small  Complete comparison of the presented keystroke biometric verification systems.}
\begin{tabular}{P{0.11\textwidth}|P{0.1\textwidth}|P{0.1\textwidth}|P{0.1\textwidth}|P{0.1\textwidth}|P{0.1\textwidth}}
\multicolumn{6}{c}{\textbf{Desktop}} \\
\bottomrule
\multicolumn{6}{c}{Global Distributions} \\
\bottomrule
\textbf{Experiment} & 
\textbf{EER (\%)$\downarrow$} & 
\makecell{\textbf{FNMR @1\%}\\ \textbf{FMR (\%)}$\downarrow$} & 
\makecell{\textbf{FNMR @10\%}\\ \textbf{FMR (\%)}$\downarrow$} &
\textbf{AUC (\%)$\uparrow$} &
\textbf{Accuracy (\%)$\uparrow$} \\
\bottomrule
TypeNet, \textit{5F} & \textbf{6.76} & \textbf{39.57} & \textbf{3.45} & \textbf{98.08} & \textbf{93.24} \\
TypeNet, \textit{4F} & 8.97 &  52.17 & 7.56 & 96.92 & 91.03 \\
TypeNet, \textit{10F} & 8.95 & 51.57 & 7.51 & 96.91 & 91.05 \\
TypeNet, \textit{11F} & 6.88 & 40.0 & 3.64 & 98.02 & 93.12 \\
\textbf{Average}  & \textbf{7.89} & \textbf{45.83} & \textbf{5.54} & \textbf{97.48} & \textbf{92.11} \\
\hline
TypeFormer, \textit{5F} & 12.95 & 75.11 & 19.03 & 94.1 & 87.05 \\
TypeFormer, \textit{4F} & 14.05 & 76.28 & 22.22 & 93.37 & 85.95 \\
TypeFormer, \textit{10F} & 12.75 & 73.51 & 18.19 & 94.31 & 87.25 \\
TypeFormer, \textit{11F} & 12.92 & 75.51 & 18.93 & 94.15 & 87.08 \\
\textbf{Average} & \textbf{13.17} & \textbf{75.1} & \textbf{19.59} & \textbf{93.98} & \textbf{86.83} \\
\bottomrule
\end{tabular}
\begin{tabular}{P{0.11\textwidth}|P{0.1\textwidth}|P{0.1\textwidth}|P{0.1\textwidth}|P{0.1\textwidth}}
\multicolumn{5}{c}{Mean Per-Subject Distributions} \\
\bottomrule
\textbf{Experiment} & \textbf{EER (\%)$\downarrow$} & \makecell{\textbf{AUC}\\  \textbf{(\%)}$\uparrow$} & 
 \makecell{\textbf{Accuracy}\\\textbf{(\%)}$\uparrow$} & \makecell{\textbf{Rank-1}\\\textbf{(\%)}$\uparrow$} \\
\bottomrule
TypeNet, \textit{5F} & \textbf{2.71} & \textbf{99.26} & \textbf{95.31} & \textbf{89.81} \\
TypeNet, \textit{4F} & 4.13 & 98.67 & 94.29 & 83.46 \\
TypeNet, \textit{10F} & 4.16 & 98.66 & 94.27 & 83.58 \\
TypeNet, \textit{11F} & 2.76 & 99.24 & 95.29 & 89.52 \\
\textbf{Average} & \textbf{3.44} & \textbf{98.96} & \textbf{94.79} & \textbf{86.59} \\
\hline
TypeFormer, \textit{5F} & 7.76 & 96.51 & 91.13 & 65.52 \\
TypeFormer, \textit{4F} & 8.62 & 95.99 & 90.4 & 62.7 \\
TypeFormer, \textit{10F} & 7.4 & 96.77 & 91.47 & 67.72 \\
TypeFormer, \textit{11F} & 7.78 & 96.48 & 91.13 & 65.07 \\
\textbf{Average} & \textbf{7.89} & \textbf{96.44} & \textbf{91.03} & \textbf{65.25} \\
\bottomrule
\end{tabular}

\begin{tabular}{P{0.11\textwidth}|P{0.1\textwidth}|P{0.1\textwidth}|P{0.1\textwidth}|P{0.1\textwidth}|P{0.1\textwidth}}
\multicolumn{6}{c}{\textbf{Mobile}} \\
\bottomrule
\multicolumn{6}{c}{Global Distributions} \\
\bottomrule
\textbf{EER (\%)$\downarrow$} & 
\makecell{\textbf{FNMR @1\%}\\ \textbf{FMR (\%)}$\downarrow$} & 
\makecell{\textbf{FNMR @10\%}\\ \textbf{FMR (\%)}$\downarrow$} &
\textbf{AUC (\%)$\uparrow$} &
\textbf{Accuracy (\%)$\uparrow$} \\
\bottomrule
TypeNet, \textit{5F} & 13.95 & 70.05 & 22.22 & 93.8 & 86.05 \\
TypeNet, \textit{4F} & 16.36 & 76.64 & 28.52 & 91.41 & 83.64 \\
TypeNet, \textit{10F} & 14.51 & 77.07 & 23.79 & 92.71 & 85.49 \\
TypeNet, \textit{11F} & 14.8 & 68.16 & 23.48 & 93.71 & 85.2 \\
\textbf{Average}  & \textbf{14.9} & \textbf{72.98} & \textbf{24.5} & \textbf{92.91} & \textbf{85.1} \\
\hline
TypeFormer, \textit{5F} & 10.17 & 70.57 & 10.46 & 95.86 & 89.83 \\
TypeFormer, \textit{4F} & 13.7 & 78.5 & 21.48 & 93.4 & 86.3 \\
TypeFormer, \textit{10F} & \textbf{9.45} & \textbf{67.67} & \textbf{8.53} & \textbf{96.22} & \textbf{90.55} \\
TypeFormer, \textit{11F} & 11.95 & 74.96 & 16.16 & 94.81 & 88.05 \\
\textbf{Average}  & \textbf{11.32} & \textbf{72.92} & \textbf{14.16} & \textbf{95.07} & \textbf{88.68} \\
\bottomrule
\end{tabular}

\begin{tabular}{P{0.11\textwidth}|P{0.1\textwidth}|P{0.1\textwidth}|P{0.1\textwidth}|P{0.1\textwidth}}
\multicolumn{5}{c}{Mean Per-Subject Distributions} \\
\bottomrule
\textbf{Experiment} & \textbf{EER (\%)$\downarrow$} & \makecell{\textbf{AUC}\\  \textbf{(\%)}$\uparrow$} & 
 \makecell{\textbf{Accuracy}\\\textbf{(\%)}$\uparrow$} & \makecell{\textbf{Rank-1}\\\textbf{(\%)}$\uparrow$} \\
\bottomrule
TypeNet, \textit{5F} & 7.99 & 96.43 & 90.89 & 68.5 \\
TypeNet, \textit{4F} & 9.64 & 95.46 & 89.5 & 65.13 \\
TypeNet, \textit{10F} & 8.3 & 96.19 & 90.64 & 66.15 \\
TypeNet, \textit{11F} & 7.99 & 96.54 & 91.03 & 68.49 \\
\textbf{Average}  & \textbf{8.48} & \textbf{96.16} & \textbf{90.51} & \textbf{67.07} \\
\hline
TypeFormer, \textit{5F} & 5.92 & 97.45 & 92.64 & 71.3 \\
TypeFormer, \textit{4F} & 8.42 & 96.09 & 90.61 & 63.1 \\
TypeFormer, \textit{10F} & \textbf{5.25} & \textbf{97.89} & \textbf{93.28} & \textbf{75.92} \\
TypeFormer, \textit{11F} & 7.41 & 96.66 & 91.42 & 66.04 \\
\textbf{Average}  & \textbf{6.75} & \textbf{97.02} & \textbf{91.99} & \textbf{69.09} \\
\bottomrule
\end{tabular}
\vspace{0.2cm}
\label{tab:all_metrics}
\end{table*}

% Overall considerations: 
% - desktop better than mobile; OK
% - Case of different threshold per subject performs better OK
% - TypeNet better in the desktop case, TypeFormer better in the mobile case 
% (TAR at 1% FAR) and rank-1 
% 66, 33 max
% - TypeNet analysis of the features
% - TypeFormer analysis of the features

\begin{figure*}[hbt!]
\begin{center}
\begin{minipage}{\textwidth}

\captionsetup[sub]{labelformat=parens}
\captionsetup{type=figure}

\begin{subfigure}{.32\textwidth}
  \includegraphics[width=\linewidth]{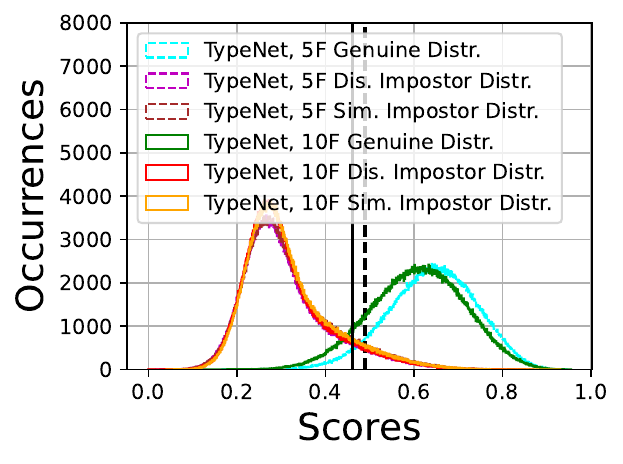}
     \caption{}
\end{subfigure}
\hfill
%\begin{subfigure}{.24\textwidth}
%  \includegraphics[width=\linewidth]{images/eer_curve_desktop.pdf}
%     \caption{}
% \end{subfigure}%
\hfill
\begin{subfigure}{.32\textwidth}
  \includegraphics[width=\linewidth]{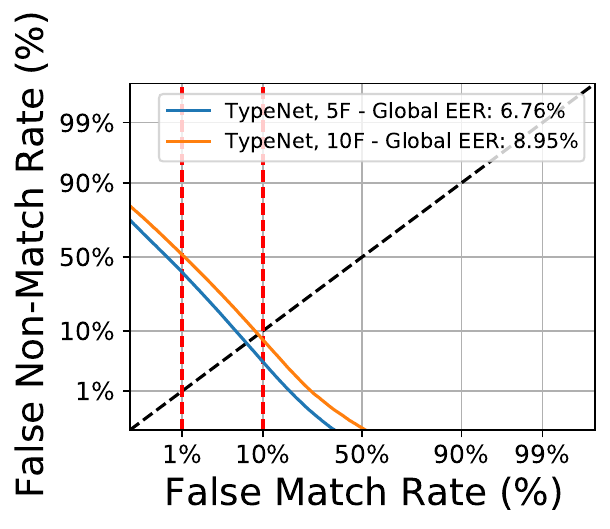}
     \caption{}
\end{subfigure}%
 \hfill
\begin{subfigure}{.32\textwidth}
  \includegraphics[width=\linewidth]{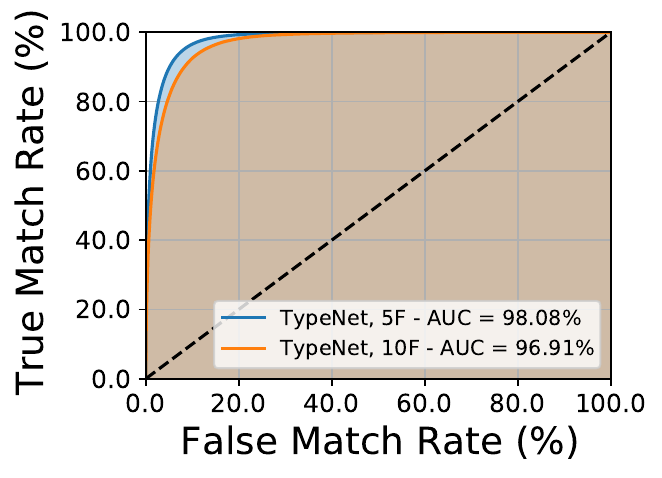}
 \caption{}
\end{subfigure}%
\hfill
\end{minipage}
\caption{The graphs show a comparison between \textbf{TypeNet} \textit{\textbf{5F}} \textbf{vs TypeNet} \textit{\textbf{10F}} in the desktop case. (a) shows the score histograms, (b) shows the DET curve, (c) shows the ROC curve.}
\label{fig:typenet_graphs}

\begin{minipage}{\textwidth}

\captionsetup[sub]{labelformat=parens}
\captionsetup{type=figure}

\begin{subfigure}{.32\textwidth}
  \includegraphics[width=\linewidth]{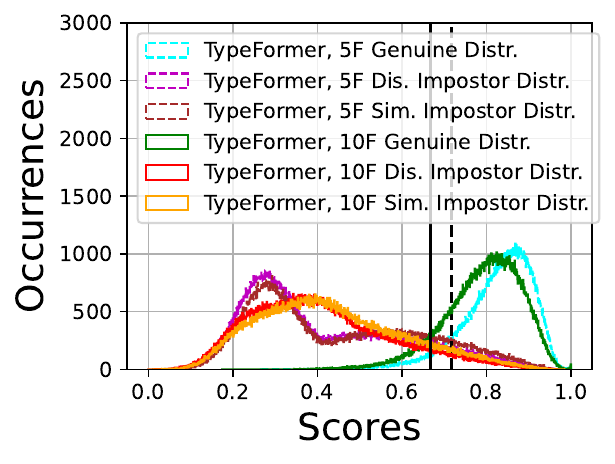}
     \caption{}
\end{subfigure}
\hfill
%\begin{subfigure}{.24\textwidth}
%  \includegraphics[width=\linewidth]{images/eer_curve_mobile.pdf}
%     \caption{}
% \end{subfigure}%
% \hfill
\begin{subfigure}{.32\textwidth}
  \includegraphics[width=\linewidth]{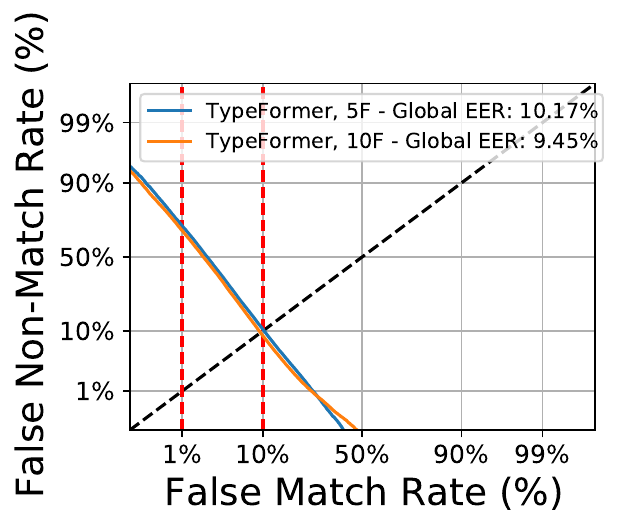}
 \caption{}
\end{subfigure}%
\hfill
\begin{subfigure}{.32\textwidth}
  \includegraphics[width=\linewidth]{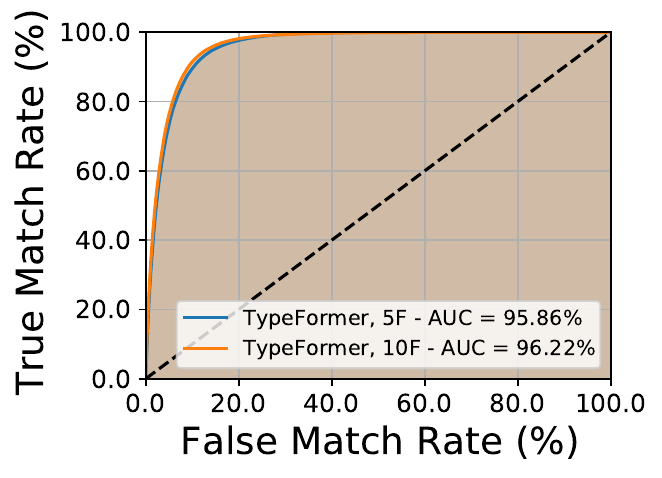}
 \caption{}
\end{subfigure}%
\hfill
\end{minipage}
\caption{The graphs show a comparison between \textbf{TypeFormer} \textit{\textbf{5F}} \textbf{vs TypeFormer} \textit{\textbf{10F}} in the mobile case. (a) shows the score histograms, (b) shows the DET curve, (c) shows the ROC curve.}
\label{fig:typeformer_graphs}

\end{center}
\end{figure*}

By observing the overall trends in the table, it is possible to notice that in the desktop case higher verification results can be achieved, possibly due to a more constrained acquisition scenario, as, in contrast to mobile devices, subjects are more likely to be sitting down and in a still position while typing on a desktop keyboard. In fact, as an example, considering both TypeNet and TypeFormer with all possible sets of features (8 experiments in total), the mean value of all EERs obtained from the global distributions in the desktop scenario is 10.53\%, whereas the corresponding value obtained in the mobile case is 13.11\%. The trend is consistent if we analyze the other metrics, such as the FNMR @1\% FMR (60.47\% vs 72.95\%) or AUC (95.73\% vs 93.99\%). 
Furthermore, the desktop case results to perform better also in the case of mean per-subject distributions. In this case, we obtain a mean EER of 5.67\% in the desktop case vs 7.61\% in the mobile case. Similarly, the mean AUC is 97.70\% vs 96.59\%, and the mean rank-1 is 75.92\% vs 68.08\%, respectively for desktop and mobile devices. 

Focusing on the four experiments involving different sets of input features, it is possible to draw some interesting conclusions. The best performing system for the desktop scenario is TypeNet, which is affected by the lack of the spatial information given by the ASCII code, and the extended set of features is not able to compensate it (6.76\% EER for TypeNet \textit{5F} vs 8.95\% EER for TypeNet \textit{10F} for global distributions). Nevertheless, despite the performance decrease, the system still shows competitive performance against the Transformer-based TypeFormer \textit{5F} or TypeFormer \textit{10F} (respectively 12.95\% EER and 12.75\% EER for global distributions). Moreover, these tendencies are consistent considering mean per-subject distributions.

The opposite outcome can be observed in the case of TypeFormer, which is the best performing system in the mobile case: the model achieved with the extended set of features (\textit{10F}) proves to be the best performing system in the mobile scenario (9.45\% EER vs second best, \textit{5F}, of 10.17\% EER for global distributions). Consequently, by introducing temporally-deepened input features, not only it has been possible to limit the performance decrease towards a more privacy-preserving verification system, but the verification performance is even significantly improved. In this case, it must be specified that the number of heads in the attention mechanism must be a multiple of the number of input features, consequently we considered (5 for \textit{5F} and \textit{10F}, 4 for \textit{4F}, and 1 for \textit{11F} due to memory constraints). In the case of TypeFormer, it is also interesting to point out that the second best performing model corresponds to the experiment \textit{5F}, with the initial set of features.

In addition, it is noticeable that subject-specific distributions lead to better verification performance in all cases. In fact, the system benefits from gaining more flexibility by setting a different threshold per subject. It is worth to highlight that this is really the case for security systems based on KD: in a real-life use case, once the system is deployed and the subject identity is verified in some other way, it is not hard to acquire multiple samples per subject. All these subject-specific data could be used as enrolment data, building a complete behavioral profile of the subject and leading to even better performance, without necessarily having to train or fine-tune the system. This would in fact be a further possible step, that leaves additional margin of improvement \cite{2018_INFFUS_MCSreview2_Fierrez}.

Another interesting observation is that it is very distinct how in the desktop case TypeNet performs better, while TypeFormer shows superiority in the mobile case. 
TypeNet is based on a two-layer LSTM RNN, while TypeFormer is based on a Transformer architecture, composed of several modules, and more parameters. 
As an example, in the desktop case the average TypeNet performance in terms of EER, FNMR @1\% FNMR, and AUC, over the four input feature experiments, is respectively 7.89\%, 45.83\%, 97.48\%, for global distributions and 3.44\% (EER), 98.96\% (AUC), 86.59\% (rank-1), for mean-per subject distributions. 
In each case, the corresponding values achieved by TypeFormer are 13.17\%, 75.10\%, 93.98\%, (global distributions), and 7.89\%, 96.44\%, 65.25\% (mean per-subject distributions), showing a significant gap. 
These trends are clearly opposite in the mobile case, where TypeFormer shows 11.32\%, 72.92\%, 95.07\% in the case of global distributions and 6.75\%, 97.02\%, 69.09\% for the mean per-subject distributions. 
In the corresponding experiments, TypeNet achieves 14.91\%, 72.98\%, 92.91\%, for global distributions and 8.48\%, 96.16\%, 67.07\% for the mean per-subject distributions. 
These trends show that LSTM RNNs seem to model well the desktop environment, while the higher variability of mobile devices is better modelled by a Transformer, which, however, does not reach the same level of performance in absolute terms.

In addition, in the case of the global distributions, also the FNMR @10\% FMR is reported. This represents a more relaxed approach, as the threshold selected is less stringent (90\% of the impostors are rejected against 99\% of FNMR @1\% FMR). According to both these metrics, TypeNet in the desktop case achieves significantly better results in comparison with all the other configurations, and its gap with TypeFormer is much greater than for the mobile case, where the two system performance is closer. These metrics are not computed for the case of mean per-subject distributions as there are not enough scores per subject for a sufficiently fine threshold resolution. They are substituted by the Rank-1 (not determinable for global distributions as subject-dependent), which also shows a more regular behavior of TypeNet in the desktop case in comparison with all other configurations.

Fig. \ref{fig:typenet_graphs} and Fig. \ref{fig:typeformer_graphs} show the curves described in Sec. \ref{subsec:curves}. In particular, it is possible to carry out a direct comparison of TypeNet in the desktop case and TypeFormer in the mobile case considering two of the four input feature sets presented above (\textit{5F} vs \textit{10F}). From left to right, the histograms of the genuine and impostor distributions are reported in Fig. \ref{fig:typenet_graphs} and \ref{fig:typeformer_graphs} (a). It is possible to see that in both rows the genuine distributions are more separated for the case of \textit{5F}, while for the mobile case (TypeFormer), the \textit{10F} setup is able to create a better separation of impostors, leading to a lower EER value. The threshold corresponding to the EER value is marked by the black line. From these macroscopic trends, the difference between the two impostor scenarios (``similar impostors'' and ``dissimilar impostors'', for comparison between subjects of the same and different demographic groups, respectively) is not very pronounced. In all graphs, the threshold values are reported in black. Then, Fig. \ref{fig:typenet_graphs} and \ref{fig:typeformer_graphs} (b) represent the DET curves (the threshold corresponding to 1\% and 10\% of FMR are respectively marked by the dashed red lines), while Fig. \ref{fig:typenet_graphs} and \ref{fig:typeformer_graphs} (c) report the ROC curves, showing similar trends from different perspectives.

\begin{table}[t]
\centering
\footnotesize 
\caption{\small Results in terms of global accuracy for the final evaluation datasets (desktop and mobile), evaluated for the different demographic groups (gender and age). STD refers to the standard deviation, whereas SER refers to the Skewed Error Rate (Sec. \ref{sec:metrics}). Both metrics are computed from all the elements of each table.}
\begin{tabular}{c|ccc}
\multicolumn{4}{c}{\textbf{TypeNet, \textit{5F}, Desktop, Global Accuracy (\%)}} \\
\bottomrule
 & \textbf{Male} & \textbf{Female} & \textbf{Average} \\   
\hline
\textbf{10 - 13} & 93.75 & 92.88 & 93.31 \Tstrut\\
\textbf{14 - 17} & 93.1 & 92.84 & 92.97 \Tstrut\\
\textbf{18 - 26} & 93.60 & 93.16 & 93.38 \Tstrut\\
\textbf{27 - 35} & 92.85 & 92.37 & 92.61 \Tstrut\\
\textbf{36 - 44} & 92.44 & 92.07 & 92.25 \Tstrut\\
\textbf{45 - 79} & 93.02 & 91.65 & 92.33 \Tstrut\Bstrut\\
\textbf{Average} & 93.13 & 92.49 & - \\
\bottomrule
\multicolumn{4}{c}{\textbf{STD}: 0.603\%, \textbf{SER}: 1.023}\\
% \bottomrule
\end{tabular}
~\\~\\~\\

\begin{tabular}{c|ccc}
\multicolumn{4}{c}{\textbf{TypeFormer, \textit{5F}, Mobile, Global Accuracy (\%)}} \\
\bottomrule
 & \textbf{Male} & \textbf{Female} & \textbf{Average} \\   
\hline
\textbf{10 - 13} & 88.63 & 87.33 & 87.98 \Tstrut\\
\textbf{14 - 17} & 89.04 & 86.52 & 97.78 \Tstrut\\
\textbf{18 - 26} & 90.69 & 88.39 & 89.54 \Tstrut\\
\textbf{27 - 35} & 89.37 & 90.05 & 89.71 \Tstrut\\
\textbf{36 - 44} & 90.1 & 90.57 & 90.34 \Tstrut\\
\textbf{45 - 79} & 90.94 & 88.43 & 89.68 \Tstrut\Bstrut\\
\textbf{Average} & 89.8 & 88.55 & - \\
\bottomrule
\multicolumn{4}{c}{\textbf{STD}: 1.38\%, \textbf{SER}:1.051}\\
% \bottomrule
\end{tabular}
% \vspace{0.2cm}
\label{tab:glob_eer_dem}
\end{table}

\begin{table*}[h!]
\scriptsize
\centering
\caption{Comparison of the presented keystroke biometric verification systems from the point of view of fairness metrics.}
\begin{tabular}{P{0.11\textwidth}|P{0.08\textwidth}|P{0.08\textwidth}|P{0.08\textwidth}|P{0.08\textwidth}|P{0.08\textwidth}|P{0.08\textwidth}|P{0.08\textwidth}}
\multicolumn{8}{c}{\textbf{Desktop}} \\
\bottomrule
\textbf{Experiment} & \makecell{\textbf{STD (\%)}$\downarrow$} & \makecell{\textbf{SER}$\downarrow$} & \textbf{FDR$\uparrow$} & \textbf{IR$\downarrow$} & \textbf{GARBE$\downarrow$}  & \textbf{SIR$_{a}$ (\%)$\downarrow$} & \textbf{SIR$_{g}$ (\%)$\downarrow$} \\
\bottomrule
TypeNet, \textit{5F} & \textbf{0.603} & \textbf{1.023} & 95.234 & 1.39 & 0.056 & 2.919 & 2.229 \\
TypeNet, \textit{4F} & 0.72 & 1.03 & 93.267 & 1.43 & 0.06 & 2.625 & 2.121 \\
TypeNet, \textit{10F} & 0.763 & 1.033 & 94.849 & 1.434 & 0.062 & \textbf{2.623} & \textbf{2.024} \\
TypeNet, \textit{11F} & 0.694 & 1.028 & 94.045 & 1.528 & 0.073 & 3.046 & 2.252 \\
\hline
TypeFormer, \textit{5F} & 0.817 & 1.034 & 97.562 & 1.296 & 0.05 & 3.137 & 2.649 \\
TypeFormer, \textit{4F} & 0.863 & 1.038 & 97.649 & \textbf{1.185} & \textbf{0.032} & 2.97 & 2.452 \\
TypeFormer, \textit{10F} & 0.847 & 1.036 & 96.71 & 1.202 & 0.037 & 2.973 & 2.493 \\
TypeFormer, \textit{11F} & 0.851 & 1.034 & \textbf{97.882} & 1.405 & 0.059 & 3.161 & 2.521 \\
\bottomrule
\end{tabular}
\begin{tabular}{P{0.11\textwidth}|P{0.08\textwidth}|P{0.08\textwidth}|P{0.08\textwidth}|P{0.08\textwidth}|P{0.08\textwidth}|P{0.08\textwidth}|P{0.08\textwidth}}
\multicolumn{8}{c}{\textbf{Mobile}} \\
\bottomrule
\textbf{Experiment} & \makecell{\textbf{STD (\%)}$\downarrow$} & \makecell{\textbf{SER}$\downarrow$} & \textbf{FDR$\uparrow$} & \textbf{IR$\downarrow$} & \textbf{GARBE$\downarrow$}  & \textbf{SIR$_{a}$ (\%)$\downarrow$} & \textbf{SIR$_{g}$ (\%)$\downarrow$} \\
\bottomrule
TypeNet, \textit{5F} & 2.234 & 1.092 & 95.727 & \textbf{2.089} & \textbf{0.099} & 6.714 & 8.305 \\
TypeNet, \textit{4F} & 1.95 & 1.077 & 94.878 & 3.413 & 0.175 & \textbf{2.883} & \textbf{3.518} \\
TypeNet, \textit{10F} & 2.037 & 1.078 & 94.948 & 2.451 & 0.17 & 4.205 & 5.15 \\
TypeNet, \textit{11F} & 1.679 & 1.074 & 95.202 & 2.36 & 0.1 & 7.076 & 8.722 \\
\hline
TypeFormer, \textit{5F} & \textbf{1.38} & \textbf{1.051} & 95.015 & 2.113 & 0.138 & 6.901 & 7.569 \\
TypeFormer, \textit{4F} & 1.764 & 1.054 & \textbf{96.042} & 2.521 & 0.142 & 6.428 & 6.262 \\
TypeFormer, \textit{10F} & 1.551 & 1.057 & 94.836 & 2.338 & 0.162 & 6.44 & 6.097 \\
TypeFormer, \textit{11F} & 1.721 & 1.075 & 93.806 & 2.247 & 0.119 & 7.601 & 7.4 \\
\bottomrule
\end{tabular}
\vspace{0.2cm}
\label{tab:demographic_metrics}
\end{table*}

\subsection{Fairness Evaluation}
Table \ref{tab:glob_eer_dem} shows the performance in terms of accuracy based on the global EER threshold (\%) considering different demographic groups. Age groups are placed along the rows, while genders along the columns. As an example, from all experiments we take into consideration TypeNet in the desktop case, and TypeFormer in the mobile case, based on the same set of features (``5F''). In both cases, it is possible to see that males achieved higher values (93.13\% against 92.49\% of global EER for TypeNet, and 89.80\% against 88.55\% for TypeFormer, respectively for males and females). It is interesting to notice that, although the STD and SER values are quite smaller in the desktop case, the difference of EER between error rates across genders is still significant. Formulating an hypothesis that would explain this trend is not immediate, and out of the scope of the current work. Furthermore, by analyzing the behavior of the systems considering different age groups, it is possible to notice that for younger subjects, TypeFormer performs worse than for older ones, while this trend is not as evident for TypeNet. Such discrepancy could be due to cultural differences related to the degree of easiness and comfort in interacting with mobile devices across different generations. 
%Moreover, while in the type of 

Table \ref{tab:demographic_metrics} shows all results of the fairness assessment provided throughout the proposed framework. Along the rows, the table is divided into two parts: the upper part presents the results in the desktop scenario, while the lower part is focused on the mobile scenario. Each half is further divided into two sections, each reporting results of one of the two models considered, TypeNet and TypeFormer, considering the four experimental configurations presented in Sec. \ref{subsec:evaluation}. Concerning the different metrics, it is necessary to point out that all metrics except $\textrm{SIR}$ are calculated considering 12 demographic groups, due to 6 age groups and 2 genders (see Sec. \ref{sec:resources}). In contrast, $\textrm{SIR}_{a}$ is computed considering scores divided by age only (square matrix of dimension 6), while $\textrm{SIR}_{g}$ considering gender only (square matrix of dimension 2), to keep the assessments of each of the two attributes independent (Sec. \ref{sec:metrics}).

By observing Table \ref{tab:demographic_metrics}, it possible to notice that there is no clear superiority of one model as in the case of the verification performance (TypeNet \textit{5F} for desktop devices, TypeFormer \textit{10F} for mobile devices). 
By carrying out an overall comparison of the desktop and mobile scenarios, it is possible to observe that demographic differentials related to age and gender tend to be higher in the mobile case. In fact, considering the mean values for each one of the metrics computed over all desktop against all mobile experiments, we obtain STD 0.77\% vs 1.79\%, SER 1.032 vs 1.07, FDR 95.90 vs 95.06, IR  1.36 vs 2.44, GARBE 0.06 vs 0.14, SIR$_{a}$ 2.80\% vs 6.03\%, SIR$_{g}$ 2.34\% vs 6.63\%. It is clear that in the desktop scenario there is an improvement in fairness according to each single metric. 

If we consider a comparison of the two architectures in each scenario, we can see that in the desktop case the average performance achieved by TypeNet in terms of SIR$_{a}$ is 2.80\% and SIR$_{g}$ is 2.16\%, while the scores of TypeFormer are respectively 3.06\% and 2.53\%, being more biased as well as less effective. A similar trend is reported for the mobile case, with TypeNet (although with lower verification performance) achieving SIR$_{a}$=5.22\% and SIR$_{g}$=6.42\%, against SIR$_{a}$=6.84\% and SIR$_{g}$=6.83\% achieved by TypeFormer.

Finally, having a look at trends due to different sets of input features within the same scenario and same model configuration, we can observe that the results are more irregular and they do not show a clear overall trend as in the previous cases. Nevertheless, considering for instance the SIR values, we can observe that generally the ASCII code is associated with higher bias in almost all experiments and scenarios, showing that comparisons among the same demographic groups are slightly harder than among different groups. %In the desktop case, for TypeNet, a slight but opposite trend is visible for metrics such as global and mean per-subject STD, suggesting that, although comparisons among the same demographic groups are slightly harder than among different groups, the system does not perform equally well within the same group for all groups.

%\FloatBarrier

\section{Conclusions and Future Work}
\label{sec:conclusions}
In this article an open experimental framework to benchmark keystroke dynamics for biometric verification is provided to the research community to alleviate the heterogeneity of the experimental protocols, metrics, and the limited size of the databases adopted in the literature. %by providing the biometric research community with a useful and significant tool to compare different solutions and classifier architectures in order to push forward state-of-the-art performance for KD. 
The framework is provided in the form of the Keystroke Verification Challenge (KVC)\footnote{\href{https://sites.google.com/view/bida-kvc/}{https://sites.google.com/view/bida-kvc/}}, hosted on CodaLab\footnote{\href{https://codalab.lisn.upsaclay.fr/competitions/14063/}{https://codalab.lisn.upsaclay.fr/competitions/14063/}}, and held within the 2023 IEEE International Conference on Big Data\footnote{\href{http://bigdataieee.org/BigData2023/}{http://bigdataieee.org/BigData2023/}}, Sorrento, Italy, December 15$^{th}$-18$^{th}$, 2023. After such term, the challenge will be made ongoing. The experimental framework is based on the two most complete and large-scale public databases of free-text keystroke dynamics up to date, collected respectively in a desktop and mobile acquisition environment, including keystroke data from more than 185,000 subjects overall. The proposed framework not only allows a complete assessment of the verification performance, but it also returns several metrics related to biometric fairness and bias based on the comparison scores, to gain new insights about KD. To this end, a novel metric, the Skewed Impostor Ratio (SIR), is proposed, designed to highlight \textit{inter}- and \textit{intra}-demographic group patterns present in the final comparison scores.

Finally, we make a first use of the proposed framework by employing two recent state-of-the-art keystroke biometric systems, TypeNet \cite{typenet} and TypeFormer \cite{typeformer}. Besides a direct comparison of the two, that shows the superiority of the former in the desktop scenario, and of the latter in the mobile one, we consider different sets of input features towards a more privacy-preserving keystroke verification system. The proposed solution is based on discarding the ASCII code, which reveals the text content, in favor of extended features in the time domain. We analyze four different experimental configurations, that focus on deepening the temporal information provided to the models in order to compensate the removal of spatial information due to the ASCII code, utilized to learn the relation in between the location of the specific key in the keyboard layout, and the correspondent time dynamics. Our experimental results show that such approach allows to maintain satisfactory performance in the desktop scenario, and even improved for mobile devices. 

Concerning future work, the next directions of research will go toward the optimisation of the model architectures to improve the recognition performance and to reduce bias. More sophisticated training approaches will also be investigated, i.e. loss functions based on the selections of hard comparisons and adaptive margins \cite{Schroff_2015_CVPR, 2022_PR_SetMargin_Morales}, specifically designed for the case of behavioral biometrics such as KD. Moreover, approaches based on the generation of synthetic subject-specific data will be considered to assess the suitability of such techniques to the problem of behavioral biometrics-based verification. To this end, KVC represents a dedicated and unified test bench for the entire biometric research community. By doing so, we aim to foster the design of innovative solutions that achieve improved performance in comparison with existing ones, that are benchmarked here.

Finally, the results of our contributed benchmark KVC in terms of demographic attribute assessment also enable further large-scale studies focused on examining the differences in subjects' typing behavior due to biological, cultural, or linguistic factors. In this sense, new findings might be of great interest for several branches of the Human-Computer Interaction (HCI) community, e.g. privacy protection \cite{delgadosantos2021survey}, security of minors online \cite{BORJ2023110039}, user experience improvement \cite{dunlop2022text2030}, etc.

% \FloatBarrier

\section*{Acknowledgment}
This project has received funding from the European Union’s
Horizon 2020 research and innovation programme under
the Marie Skłodowska-Curie grant agreement No. 860315 (PriMa project), INTER-ACTION project (PID2021-126521OB-I00
MICINN/FEDER) and HumanCAIC project (TED2021-131787BI00 MICINN).

 % the bibliography file.
\bibliographystyle{unsrt}
\bibliography{0_Main}

\end{document}